\documentclass[11pt,letterpaper]{article}
\usepackage{acl2012}
\usepackage{times}
\usepackage{latexsym}
\usepackage{multirow}
\usepackage{float}
\usepackage{amsmath,amsfonts,amssymb}
\usepackage{algorithm}
\usepackage[noend]{algpseudocode}
\usepackage{comment}
\usepackage{multirow}
\makeatletter
\def\BState{\State\hskip-\ALG@thistlm}
\makeatother

\usepackage{float}
\usepackage{mathtools}
\usepackage{url}
\newcommand\norm[1]{\left\lVert#1\right\rVert}
\newcommand\tab[1][1cm]{\hspace*{#1}}

\usepackage[textsize=scriptsize]{todonotes}
\RequirePackage{enumitem}
\setlist{nolistsep}
\usepackage{titlesec}
\titlespacing*{\section}{0pt}{1.3ex plus .1ex minus .1ex}{.8ex plus .1ex}
\titlespacing*{\subsection}{0pt}{1.1ex plus .1ex minus .1ex}{.6ex plus .1ex}
\titlespacing*{\paragraph}{0pt}{.1ex plus .1ex minus .1ex}{.3em}

\newcommand{\bi}{\begin{itemize}}
\newcommand{\ei}{\end{itemize}}
\newcommand{\BE}{\begin{enumerate}}
\newcommand{\EE}{\end{enumerate}}

\newcommand{\etal}{\mbox{\it et al.}}

\newcommand{\initab}{                           
\begin{tabbing}
XXX \= XXXX \= \kill
}
\newcommand{\begpub}{
\begin{quotation}
\noindent
}

\newcommand{\finpub}{
\end{quotation}
}

\title{Joint Matrix-Tensor Factorization for Knowledge Base Inference}

\author{
Prachi Jain\textsuperscript{1}, Shikhar Murty\textsuperscript{1}\thanks{First two authors contributed equally to the paper} , Mausam\textsuperscript{1}, and Soumen Chakrabarti\textsuperscript{2}\\
\textsuperscript{1}Indian Institute of Technology Delhi\\
\textsuperscript{2}Indian Institute of Technology Bombay}

\date{}
\newcommand{\es}{e_1}
\newcommand{\eo}{e_2}
\newcommand{\ep}{ep_{12}}
\newcommand{\ves}{\vec{\es}}
\newcommand{\veo}{\vec{\eo}}
\newcommand{\vep}{\vec{ep}_{12}}
\newcommand{\vr}{\vec{r}}
\renewcommand{\etal}{\emph{et~al.}}

\begin{document}
\maketitle
\begin{abstract}
While several matrix factorization (MF) and tensor factorization (TF) models have been proposed for knowledge base (KB) inference, they have rarely been compared across various datasets. Is there a single model that performs well across datasets? If not, what characteristics of a dataset determine the performance of MF and TF models? Is there a joint TF+MF model that performs robustly on all datasets? We perform an extensive evaluation to compare popular KB inference models across popular datasets in the literature. In addition to answering the questions above, we remove a limitation in the standard evaluation protocol for MF models, propose an extension to MF models so that they can better handle out-of-vocabulary (OOV) entity pairs, and develop a novel combination of TF and MF models. We also analyze and explain the results based on models and dataset characteristics. Our best model is robust, and obtains strong results across all datasets.
\end{abstract}

\section{Introduction}

Inference over knowledge bases (KBs) has received significant attention within NLP research in the last decade. Most of the early works on this task focus on adapting probabilistic formalisms such as Markov Logic Networks and Bayesian Logic Programs for inferring new KB facts \cite{schoenmackers2008scaling,niu2012elementary,raghavan2012learning}. The formalisms require a set of inference rules as input, which can be generated automatically using statistical regularities in KBs \cite{schoenmackers2010learning,berant2011global,nakashole2012patty,DBLP:conf/naacl/JainM16}. 

Recent research on this task has integrated the two components of rule learning and fact inference into one joint deep learning framework. This eschews explicit representation and learning of inference rules, and instead employs a way to score a (possibly new) KB fact $(\es,r,\eo)$ directly. Various algorithms differ in their scoring functions, which score a KB fact using different model assumptions.

This line of research can be further subdivided into two broad categories: matrix factorization and tensor factorization . In both cases the models learn one or more embeddings of the relation $r$, however, they differ in their treatment of entities $\es$ and $\eo$. Tensor factorization (TF) approaches (e.g., E \cite{reidel-naacl13}, TransE \cite{garciaduran2015translating}, DistMult \cite{yang15export:241703}, Rescal \cite{nickel2011three} models) learn independent embeddings for $\es$ and $\eo$, whereas matrix factorization (MF) methods (e.g., F \cite{reidel-naacl13} model) learn an embedding per entity-pair $(\es,\eo)$. Except for one paper making some early progress \cite{singh2015towards}, their relative benefits have not been studied in detail.

More importantly, MF and TF have been rarely compared on the same datasets. In particular, three popular KBs are commonly used for TF research (WN18, FB15K, FB15K-237) and one for MF research (NYT+FB, New York Times articles annotated with Freebase entities), but rarely has a model been tested on all four. To the best of our knowledge, no paper reports the performance of E and F models on WN18 or FB15K, TransE on FB15K-237 or NYT+FB, and DistMult on NYT+FB.


\paragraph{Contributions:}
We unify several closely related tasks into KB inference (KBI) from a combination of incomplete KBs and text corpus. Our goal is to design inference algorithms that work robustly across diverse input combinations and datasets.

To that end, we first compare E, TransE, F and DistMult (DM) models on all four datasets.
The comparison reveals that subtle issues arise in the design of training and evaluation procedures when TF methods are compared against or combined with MF methods. Special care is needed to handle out-of-vocabulary (OOV) entity-pairs during evaluation. Otherwise an MF algorithm may appear to perform better than it really does, as in the case of F's performance on FB15K-237 \cite{toutanova2015representing}. 

In response, we present the first unified KBI evaluation protocol that can meaningfully compare MF and DM approaches across several datasets. F's performance deteriorates using the KBI evaluation protocol.
The main reason is an {\em ad hoc} handling of OOV entity-pairs by F.  We then propose an enhancement of F that explicitly learns OOV entity-pair vectors.  This significantly improves F's performance, but DistMult (DM) remains the most robust solution across all datasets.

Further analysis shows that datasets associated with TF approaches have high OOV-rate in most test folds, naturally resulting in F performing poorly.  However, F performs well on the dataset with low OOV rate.  Our final contribution is a robust joint algorithm combining DM and F, which is competitive with both models on all datasets, and also outperforms the joint models proposed earlier.

Along with the above results, we contribute open-source implementations\footnote{\em https://github.com/dair-iitd/kbi} of all the methods and testing protocols investigated.

\section{Background and Experimental Setup}
\label{sec:back}

We propose knowledge base inference (KBI) as a task that unifies several closely related tasks in prior work, particularly, knowledge base completion (KBC), link prediction, and relation extraction (RE). In KBC and link prediction, new tuples are inferred from an incomplete structured KB. In RE, relations are inferred between entities mentioned in an unstructured corpus. It is natural \cite{toutanova2015representing} to unify these paradigms, along with textual tuples from OpenIE \cite{etzioni2011open}.  

Specifically, we are given an incomplete KB that consists of a set of entities $\mathcal{E}$ and relations $\mathcal{R}$. $\mathcal{R}$ may contain only semantic relations, only textual relations or a combination of both, as we want inference to benefit from structural regularities among unnormalized and canonical relations, even if these are not reconciled.  The KB also contains $\mathcal{T}$, a set of known valid tuples $t\in\mathcal{T}$. A tuple $t=\langle \es, r, \eo \rangle$ consists of a subject entity $\es \in \mathcal{E}$, object entity $\eo \in \mathcal{E}$, and relation $r \in \mathcal{R}$. We use a shorthand $\ep$ to refer to entity pair $(\es,\eo)$. Our goal is to predict the validity of any new tuple not present in the KB.

Our focus is on the numerous neural models $M$ that learn distributed representations (embedding vectors $\in \mathbb{R}^d$) of entities and relations. At a high level, each model defines a way to compute a {\em score} for the tuple $\langle \es, r, \eo \rangle$ based on some factorization. There are two broad categories of factorization models --- tensor factorization (TF) and matrix factorization (MF). Both these kinds of models learn one (or more) embedding of $r$ denoted by $\vr$. However, they differ in their treatment of entities. TF models learn embeddings for each entity $\ves$ and $\veo$, whereas MF models learn a single embedding for each entity pair $\vep$.\footnote{Some models may also learn matrix embeddings instead of vectors \cite{nickel2011three,socher2013reasoning}. We don't study these, as they are typically outperformed by the models implemented in this paper \cite{yang15export:241703,trouillon2016complex}.}

Different models differ primarily in the function $\phi^M(e_1, r, e_2)$ that combines these embeddings to score a tuple. A higher value of $\phi^M$ denotes a model's higher confidence that the tuple is valid. Table \ref{model_scoring_function} lists the scoring functions used by four popular models, which are the focus of our paper. These are E, F \cite{reidel-naacl13}, TransE \cite{garciaduran2015translating}, and DistMult \cite{yang15export:241703}. Of these, F is an MF model, since it uses the $\vep$ embeddings, while the rest three are TF models. Note that E learns two embedding vectors $\vec{r_s}$ and $\vec{r_o}$ for a relation $r$. DM uses an element-wise multiplication $\bullet$ in its scoring function.

Our choice of these models is guided by the fact that these algorithms either form the basis of several recent papers on KB inference or are popular baselines for comparison studies \cite{toutanova2015representing,trouillon2016complex,demeesterregularizing,rocktaschel-naacl15,verga2016generalizing,verga2015multilingual,singh2015towards}.


\begin{table}[th]
\centering
\begin{tabular}{|l|l|}
\hline
\textbf{Model ($M$)} & \textbf{Scoring function ($\phi^M(e_1,r,e_2)$)} \\ \hline
TransE & $-\norm{\ves + \vr - \veo}_2$ \\ 
F & $\vr^\top \cdot \vep$ \\ 
E & $(\ves^\top \cdot \vec{r_s}) + (\veo^\top \cdot \vec{r_o})$ \\ 
DistMult & $\vec{r}^\top \cdot (\vec{\es} \bullet \vec{\eo})$ \\ \hline
\end{tabular}
\caption{Scoring functions for various models. Larger value implies more confidence 
in the validity of the triple. `$\cdot$' denotes dot product and
`$\bullet$' denotes element-wise multiplication.} \label{model_scoring_function}
\end{table}




\paragraph{Loss functions:} The models are trained such that tuples observed in the KB have higher scores than unobserved ones. Several loss functions have been proposed; we implement two common ones in this work: log-likelihood based loss and max margin loss. Both loss functions sample a negative set $Neg(\es, r)$ for every tuple, computed as $\{\langle e_1,r,e_2' \rangle|e_2'\in {\cal E} \wedge \langle \es, r, e_2' \rangle\notin \mathcal{T}\}$, i.e., tuples formed by uniformly sampling entities that are not apriori known to be valid. Similarly, the set $Neg(r, \eo)$ is sampled. 


To define a log-likelihood based loss for $M$,
Toutanova \etal~\shortcite{toutanova2015representing}
first model an approximate\footnote{For a rigorous estimate, we need to include the numerator also in the denominator, and correct the denominator by the ratio of population to sample size.} conditional probability:
\begin{multline}
p^M(\eo | r, \es; \theta) = \\
\frac{\exp(\phi^M(\es,r,\eo; \theta))}{\sum\nolimits_{\langle \es,r,e_{2}' \rangle \in Neg(\es,r)} \exp(\phi^M(\es,r,e_{2}'; \theta))}
\end{multline}
Here $\theta$ represents model parameters: the embeddings for each relation and entity (or entity pair). $p^M(\es | r, \eo; \theta)$ is estimated similarly using $Neg(r, \eo)$. The log-likelihood loss to minimize is
\begin{multline}
\mathcal{L}_{ll}^M(\mathcal{T},\theta)=-\left[\textstyle 
\sum_{\langle e_1,r,e_2 \rangle \in\mathcal{T}} \log p^M(e_1|r,e_2;\theta) \right. \\ \left.
+ \textstyle \sum_{\langle e_1,r,e_2 \rangle \in\mathcal{T}} \log p^M(e_2|r,e_1;\theta)\right]
\end{multline}
On the other hand, max-margin loss minimizes a margin-based ranking criterion \cite{garciaduran2015translating}:
{\small 
\begin{equation}
\!\!\mathcal{L}_{mm}^M(\mathcal{T}, \theta)=
\sum_{t \in \mathcal{T}}\!\sum_{t' \in Neg(t)}\!\!\left[\gamma+\phi^M(t')-\phi^M(t) \right]_+
\end{equation}
}
where $t=\langle \es,r,\eo \rangle$, $Neg(t)=Neg(\es,r)\cup Neg(r,\eo)$, $\gamma$ is the margin and $[x]_+=\max\{0,x\}$.

Finally, note that since MF models operate over entity pairs, they do not need two $Neg$ sets. They use one set where new entity pairs $(\es',\eo')$ are sampled such that $\langle \es',r,\eo' \rangle \notin\mathcal{T}$. These negative entity pairs are sampled only from the entity pairs found in $\mathcal{T}$, since embeddings for only those pairs get learned. 



\paragraph{MF vs.\ TF Models:} Limited comparisons have been made between the MF and TF families. Toutanova et al.\ \shortcite{toutanova2015representing} compare F with some TF models on one dataset and find that F does not perform as well as TF. Singh et al.\ \shortcite{singh2015towards} use a series of artificial experiments to conclude that MF models typically perform well on tasks where there is significant relation synonymy in the data, whereas TF models perform better when there are latent types for each relation that need to be predicted. Singh and Toutanova experiment on one real dataset each and show the value of (different) joint MF-TF models on those datasets. We revisit these in Section \ref{sec:hybrid}.

\subsection{Datasets}

Most KB inference systems have used one or more of four popular KBs for evaluation. These include WN18 (eighteen Wordnet relations \cite{garciaduran2015translating}) and three datasets over Freebase (FB). One dataset is FB15K \cite{garciaduran2015translating} that has 1,345 relations. Another dataset is FB15K-237, which is a subset of FB15K comprising 237 relations that seldom overlap in terms of entity pairs~\cite{toutanova2015representing}. The fourth dataset is NYT+FB, which, along with FB triples, also includes dependency path-based textual relations from New York Times, the mentions of entities in which are aligned with entities in Freebase~\cite{reidel-naacl13}.

Our literature search reveals that no algorithm has been tested on all datasets. To the best of our knowledge, no paper reports results of E and F models on WN18 or FB-15K, TransE on FB15K-237 or NYT+FB, and DistMult on NYT+FB. To better understand the strengths and weaknesses of each model (especially TF vs. MF), we compare all models on all datasets. We also release their open source implementations for further research.

\subsection{Standard Evaluation Protocol}
\label{sec:evalprot}

Since we wish to run these experiments at scale, we follow one of the common evaluation protocols that can be run completely automatically. This method splits the KB into train ($\mathcal{T}_{tr}$) and test tuples ($\mathcal{T}_{ts}$). The system can access only $\mathcal{T}_{tr}$ during training. For each test tuple, $ \langle e_1^*, r^*, e_2^* \rangle \in \mathcal{T}_{ts}$, a query $\langle e_1^*, r^*, ? \rangle$ is issued to the trained model $M$. The model then ranks all entities $e_2\in\mathcal{E}$ by decreasing $\phi^M(e_1^*, r^*, e_2)$. A higher rank of $e_2^*$ in this list suggests a better performance of the model. The metrics used to compare two algorithms are mean reciprocal rank (MRR) and the percentage of $e_2^*$s obtained in top 10 results (HITS@10). 

The testing procedure is typically run with two modifications. First, it is possible that some of the $e_2$s ranked higher than $e_2^*$ may form known valid tuples $\langle e_1^*, r^*, e_2 \rangle$ --- it is unfair to penalize the model for predicting these. The {\em filtered} metrics remove the set $\{e_2| \langle e_1^*, r^*, e_2 \rangle \in \mathcal{T}_{tr}\cup\mathcal{T}_{ts}\}$ from the ranked list \cite{garciaduran2015translating}.

The second modification applies primarily to MF models. In MF, an embedding is learned only for entity pairs that appear in $\mathcal{T}_{tr}$. Therefore, it is futile to score every $ \langle e_1^*, r^*, e_2 \rangle$ over a large range of $e_2$s, for most of which, $\vec{ep}_{12}$ is not even known. Instead, only those $e_2$s in a smaller set
\begin{align}
E_2 &=\{e_2|\exists r: \langle e_1^*,r,e_2 \rangle \in\mathcal{T}_{tr}\cup \mathcal{T}_{ts}\}
\label{eq:wrongE2}
\end{align}
are considered as candidates for ranking \cite{toutanova2015representing,verga2016generalizing}. If entity pair $(e_1^*,e_2)$ is not trained then a random vector is assumed for $\vec{ep}_{1^*2}$.


\begin{table*}[th]
\centering
{\small
\begin{tabular}{|l|cc|cc|cc|cc|} \hline
\multirow{2}{*}{Model} & \multicolumn{2}{c|}{FB15K} & \multicolumn{2}{|c|}{FB15K-237} & \multicolumn{2}{c|}{WN18} & \multicolumn{2}{c|}{NYT+FB} \\
& MRR & HITS@10 & MRR & HITS@10 & MRR & HITS@10 & MRR & HITS@10 \\ \hline
\hline
DistMult & {\bf 44.70} & 66.26 & {\bf 34.07} & {\bf 52.93} & {\bf 75.91} & {\bf 94.12} & 62.48 & 72.17 \\ 
E & 22.38 & 34.56 & 30.71 & 44.84 & 2.36 & 4.78 & 7.81 & 19.14 \\ 
TransE & 43.11 & {\bf 71.97} & 1.88 & 0.01 & 37.15 & 84.96 & 7.98 & 44.05 \\ 
F  & 33.62 & 60.20 & 28.01 & 64.76 & 82.95 & 98.84 & 89.28 & 97.84 \\ \hline
\multicolumn{9}{c}{} \\ \hline
F (KBI eval) & 13.35 & 17.03 & 0.0 & 0.0 & 0.14 & 0.20 & 74.34 & 80.01 \\ \hline
\multicolumn{9}{c}{} \\ \hline
MFreq$(e_2|r^*)$ & 24.91 & 36.03 & 33.05 & 47.60 & 3.10 & 5.28 & 0.90 & 1.56 \\ 
MFreq$(e_2|e_1^*)$ & 8.22 & 15.61 & 0.01 & 0.01 & 0.00 & 0.00 & {\bf 79.34} & {\bf 94.93} \\ \hline
\end{tabular}
\caption{The first four rows compare four models on four datasets using the standard evaluation protocol. The fifth row shows F's performance using our proposed KBI evaluation protocol. The last two rows reports results of two most-frequent sanity-check baselines.}
\label{eval-scores}
}
\end{table*}

\section{Comparison under Standard and Unified KBI Evaluation Protocols}

\subsection{Training Details}

We first re-implement all algorithms in a common framework written using Keras/Theano \cite{chollet2015keras,2016arXiv160502688short}. We use 100 dimensional vectors for all models. They are trained using mini-batch stochastic gradient descent with AdaGrad on K40 GPUs with a learning rate of 0.5. We pre-compute 200 negative samples per tuple. We set margin $\gamma$ to 1 for max margin loss. Following previous work \cite{yang15export:241703} all entity and entity-pair vectors are re-normalized to have a unit norm after each batch update. We use a batch size of 20,000 for training. We train all models for 200 epochs. We use early stopping on validation set (a small subset of training set), to prevent our models from overfitting.


We train each model on each dataset using both log-likelihood (LL) and max-margin (MM) loss functions. We pick the best loss function for every setting. In particular, we find that TransE performs much better with MM loss. LL loss works better or at par in all other models except that MM outperforms LL for DistMult on WN18 dataset. 

\begin{table*}[t]
{\small
\begin{minipage}{0.5\linewidth}
\centering
\begin{tabular}{|l|c|c|}
\hline
$\langle$ Bill Gates, lives in, ?$\rangle$ & F (old) & F (new) \\ \hline
(Bill Gates, lives in, Seattle)                          & 5.34 & 5.34                       \\
{\bf \textit{(Bill Gates, lives in, Medina)}}                                  & 0.04 & -1.4                       \\
\textit{(Bill Gates, lives in, New York)} & ? & -1.4 \\
\textit{$\mathrel{\makebox[\widthof{(Bill Gates, }]{\vdots}} \mathrel{\makebox[\widthof{lives in, }]{\vdots}} \mathrel{\makebox[\widthof{New York)}]{\vdots}}$} & ? & -1.4
 \\ \hline
\hline
Reciprocal rank & 0.5 & $\sim$0.0          \\ \hline
\end{tabular}
\\ 
\end{minipage}%
\begin{minipage}{0.5\linewidth}
\centering
\begin{tabular}{|l|c|c|}
\hline
$\langle$ Tina Fey, lives in,  ?$\rangle$ &  F (old) &  F (new) \\ \hline
\bf {(Tina Fey, lives in, New York)} & 2.30 & 2.30                       \\
(Tina Fey, lives in, Seattle) & 1.1 & 1.1               \\
{\em (Tina Fey, lives in, Medina)} & ? & -2.12                       \\
\textit{$\mathrel{\makebox[\widthof{Tiny Fey, }]{\vdots}} \mathrel{\makebox[\widthof{lives in, }]{\vdots}} \mathrel{\makebox[\widthof{Medina}]{\vdots}}$} & ? & -2.12\\
\hline
 \hline
Reciprocal rank & 1 & 1  \\ \hline
\end{tabular}
\\ 
\end{minipage}
\caption{Original F with old evaluation protocol vs. F (trained OOV vector) with KBI evaluation protocol. Bold means the gold tuple, and italics means that entity-pair isn't seen in training. (a) Bill Gates is seen with one $e_2$ in training -- not the gold answer, (b) Tina Fey is seen with two $e_2$s including the gold answer.} 
\label{eval-example}
}
\end{table*}

We follow the train-dev-test splits used in previous experiments for FB15K, WN18, and FB15K-237. The testsets $\mathcal{T}_{ts}$ are 3--10\% random samples from $\mathcal{T}$. For NYT+FB, previous works had experimented on a test fold with only 80 correct tuples \cite{reidel-naacl13}. Since such a test set is rather small, and in keeping with our other data sets, we create our own train-test splits by randomly sampling about 2\% tuples from $\mathcal{T}$. Only tuples with FB relations are used in the test set similar to previous experiments on this dataset.

\subsection{Preliminary Results}

The first four rows of Table \ref{eval-scores} report the performance of all the models across the datasets. We observe DistMult (DM) to be an overall winner among tensor factorization models -- E has good performance on FB15K-237, whereas TransE gets good scores on FB15K, however DM emerges the most robust. For TF models on three datasets (FB15K, FB15K-237, WN18) our experiments are able to replicate (or improve upon) various results reported in prior works \cite{yang15export:241703,garciaduran2015translating,toutanova2015representing}.\footnote{\cite{yang15export:241703} report a higher MRR for DM on WN18.} Since NYT+FB is a new test split, and F hasn't been tested on other datasets, those results can't be directly compared against previous work.

We also find that F outperforms DM on two datasets by wide margins and doesn't perform as well as DM on the other two. It appears that a qualitative analysis of DM vs. F will shed light on their relative strengths and weaknesses. Our analysis reveals a limitation in the standard evaluation protocol that can inflate F's performance scores for OOV entity pairs.

\subsection{KBI Evaluation Protocol}

Recall the second modification from Section \ref{sec:evalprot}. When ranking possible entities $e_2$ using the score $\phi(e_1^*, r^*, e_2)$ from MF models, the standard evaluation protocol operates over a subset $E_2$, instead of all entities in $\mathcal{E}$. This is because many entity pair embeddings $(e_1^*, e_2)$ are not even trained in the model, and hence their scores will be meaningless. We call these OOV entity pairs. $E_2$ contains all entities for which the entity pair $(e_1^*, e_2)$ is trained. But, additionally, all such $e_2^*$s are added to $E_2$ that are gold entities for some query $\langle e_1^*, r^*, ? \rangle$ in test set. If these are not trained, a random vector is assumed for them.

Table \ref{eval-example}(a) illustrates an extreme case where the gold entity pair (Bill Gates, Medina) is not seen in training, and only one $e_2$ (Seattle) is seen with $e_1^*$. Here, the MRR for F model will be computed as 0.5 --- a gross overestimation! Implicitly, $(e_1^*, e_2^*)$ is getting ranked higher than all other OOV $(e_1^*, e_2)$s, whereas they should all be equal. In other words, the mere presence of $\mathcal{T}_{ts}$ in Eqn \eqref{eq:wrongE2} leaks information.

Ideally, an evaluation protocol for KBI, that is tolerant to OOV entity pairs, must assume all OOV entities at the same rank and output the average value over all possible rankings for them. In our enhanced protocol, we assume one random OOV entity pair vector $(e_1^*, e_{oov})$, identify {\em all} $e_2\in\mathcal{E}$ that are OOV, assign them all the same score from the model and compute aggregate scores based on all possible rankings of such OOV entities. In our example of Table \ref{eval-example}(a), the MRR will be computed as the average of $\frac{1}{2}, \frac{1}{3}$, \dots which is a very small number.


We note that most existing MF models have been tested on test splits in which none of the gold entity pairs are OOV (except FB15k-237). Hence, the results reported in most previous papers are not affected by our proposed fix. Even otherwise, if variants of MF models are being compared among themselves, while they may overestimate performance somewhat, the relative ordering of various models may not be affected. On the other hand, OOVs become a central issue when MF models are compared against or combined with TF models, since realistic levels of sparsity are very different in the two models. We elaborate on this below. 

\subsection{Results Adjusted for KBI Evaluation}

When the KBI evaluation protocol is used, F's performance on all datasets drops drastically, to the extent that its performance is practically zero on two datasets, and extremely weak on the third. However, it continues to have the best numbers for NYT+FB.  Our evaluation sanitizes the published numbers for F on FB15K-237 \cite{toutanova2015representing}. 

\begin{table}[h]
\centering
{\small
\begin{tabular}{|l|c|c|c|}
\hline
\textbf{Dataset} & \textbf{$|\mathcal{E}|$} & \textbf{$|\mathcal{R}|$} & \textbf{{ep} OOV (\%)} \\ \hline
FB15K            &14,951  & 1,345   & 68.70            \\ 
FB15K-237        &14,541  & 237    & 100.00              \\ 
WN18             & 40,943 & 18     & 99.52            \\ 
NYT+FB           & 24,528 & 4,111   & 0.75             \\ \hline
\end{tabular}
\caption{No. of distinct entities, no. of relations and entity pair OOV rate, i.e., percentage of tuples in test set, whose entity pairs weren't seen while training.}
\label{dataset-oovs}
}
\end{table}

Why is there such a significant drop in F's scores? The answer lies in entity pair OOV rates for these datasets, i.e., the percentage of tuples in test set whose entity pairs were not seen while training. Table \ref{dataset-oovs} reports some statistics about the datasets as well as their test sets. We notice that FB15K, FB15K-237 and WN18 all have a very high OOV rates, which is strongly correlated with poor performance of F. On the other hand, NYT+FB has a tiny OOV rate and F performs well on it.

Indeed, it is obvious that if the gold entity pair is not even seen while training, an MF model won't be able to predict it, since it learns each entity-pair vector separately. On the other hand, a TF model, by virtue of learning each entity vector separately (\emph{single entity} OOVs are very infrequent in these datasets), could combine its knowledge of each individual entity for predicting unseen entity-pairs.  Singh et al.\ \shortcite{singh2015towards} contribute some theoretical differences between MF and TF models (see Section~\ref{sec:back}). Our analysis on the basis of entity-pair OOVs adds to that understanding. Moreover, we believe that OOVs, and more generally, data sparsity, offer a more practical insight into differences between two model types --- representation in MF necessitates more data points per entity pair, whereas TF is more robust to sparse datasets. 

Why does DM model perform the best? While we do not have a conclusive answer to this question, we believe that two reasons could act in DM's favor. First, like F, DM also has a representation of an entity pair. However, rather than associating an opaque single vector with each entity pair (where the role of individual entities cannot be identified), DM \emph{composes} the entity-pair vector using entity vectors, as $\vec e_1 \bullet \vec e_2$. Thus, it is likely able to exploit some power of matrix factorization, while still being robust to data sparsity. Secondly, even TransE can be seen as composing an entity-pair vector ($\ves-\veo$), but it is additive, whereas DM is multiplicative. Previous work on word vectors has shown that multiplicative scores often outperform additive ones  as they amplify smaller differences and reduce larger ones \cite{levy14linguistic,stanvosky15open}.

\begin{table*}[th]
\centering
{\small
\begin{tabular}{|l|cc|cc|cc|} \hline
\multirow{2}{*}{Model} & \multicolumn{2}{c|}{FB15K} & \multicolumn{2}{c|}{WN18} & \multicolumn{2}{c|}{NYT+FB} \\
& MRR & HITS@10 & MRR & HITS@10 & MRR & HITS@10 \\ \hline
\hline
F (random) & 13.35 & 17.03 & 0.14 & 0.20 & 74.34 & 80.01 \\ 
F (average) & \textbf{18.27} & \textbf{24.62} & 0.13 & 0.16 & 71.65 & 76.80 \\ 
F (trained) & 17.94 & 23.82 & \textbf{0.19} & \textbf{0.24} & \textbf{81.51} & \textbf{93.67} \\ \hline
\end{tabular}
\caption{Results on F model after explicitly modeling OOV vectors. OOV training outperforms other baselines, especially for NYT+FB. Results on FB15k-237 not reported, due to 100\% entity pair OOV rate.}
\label{oov-train}
}
\end{table*}

\subsection{Most-Frequent Baselines}

To improve our understanding of the difficulty of each dataset and the quality of each model, we introduce two baselines for our task. Given a query, $\langle \es^*, r^*, ? \rangle$ our first baseline ranks all entities based on the frequency of their occurrence with relation $r^*$, i.e., it orders each entity $e_2$ based on the cardinality of the set $\{t|t=\langle e_1,r^*,e_2 \rangle \wedge t\in\mathcal{T}_{tr}\}$. A similar baseline orders each entity $e_2$ based on its frequency of occurence with $e_1^*$, i.e., based on cardinality of the set $\{t|t=\langle e_1^*,r,e_2 \rangle \wedge t\in\mathcal{T}_{tr}\}$. We name these baselines MFreq$(e_2|r^*)$ and MFreq$(e_2|e_1^*)$ respectively. Our motivation to introduce these is to check whether existing models are able to learn beyond such simple baselines or not.

The last two rows of Table \ref{eval-scores} report the performance of these baselines. It is satisfying to see that for FB15K and WN18 datasets, DM outperforms the baselines by large margins. However, for FB15K-237, DM is only marginally better than MFreq$(e_2|r^*)$. A closer analysis reveals that this dataset is constructed so that there is minimal entity-pair overlap between relations. Thus, how would any model predict the best $e_2$ for a query $\langle e_1^*, r^*, ? \rangle $? If entity pairs haven't been repeated much, a natural approach may just find the most frequent entities seen with the relation and order based on frequency. We checked some high MRR predictions made by DM and found that often questions like, what is the language of a specific website were answered correctly as English. This is likely not because DM figured out the language of each website, but because English was the most frequent one.

We also observe that E's performance remains broadly similar to the performance of MFreq$(e_2|r^*)$. We attribute this to E's scoring function, since given $e_1^*$ and $r^*$, the only term relevant for ranking $e_2$s is $\veo^\top\cdot\vr_o$, i.e., the model looks for compatibility with $r^*$ and ignores $e_1^*$ completely. 

Finally, for NYT+FB, MFreq$(e_2|e_1^*)$ beats F model significantly suggesting that while F is the best model on that dataset, it is not good enough. We explore this further in the next section.

\section{OOV Training for KB Inference}

The previous section highlights the importance of OOV entity-pairs in the performance of MF models. In general, a robust model must gracefully handle unseen entities/entity-pairs. A natural extension is to explicitly model an OOV entity-pair vector for F model (and OOV entity vector for TF models). In particular, we represent a vector $(e_{oov}, e_{oov})$ vector for F and $e_{oov}$ for TF.\footnote{We also tried learning several entity pair OOV vectors of the form $(e_1, e_{oov})$, but that didn't give us a better performance.} This modification means that OOV entity-pairs will have the same score. 


OOV vectors can be trained in many ways. We develop two baselines that don't train the vectors explicitly. One baseline assigns a {\em random} value to $(e_{oov},e_{oov})$. Another is an {\em average} baseline that computes $(e_{oov},e_{oov})$ as the average of the vectors of all $(e_1, e_2)$ pairs that occur only once in training.

We also propose a procedure to {\em train} the OOV vectors. The high-level motivation is that we wish to score a known tuple higher than a tuple with an OOV. To ensure this, we add $(e_{oov}, e_{oov})$ in the $Neg$ set for each train tuple. This encourages the model to learn embeddings such that $\phi^F(e_1, r, e_2) > \phi^F(e_{oov}, r, e_{oov})$.  Thus, we ensure that the performance of F is maintained when the gold entity pair is seen in training. Table \ref{eval-example}(b) illustrates an example where the correct answer (New York) is seen with Tina Fey and OOV training doesn't displace its position. For a TF model, we follow an analogous procedure to train an OOV vector $\vec{e}_{oov}$.

\noindent{\bf Results: } Since the fractions of OOV entities ($e_2^*$s) in the testsets are rather small, OOV training doesn't benefit TF models much. However, it makes substantial improvements in F's performance. Table \ref{oov-train} compares trained OOV embeddings to the two baselines for F. We find that training of OOVs overall performs better (or at par) with averaging baseline. F's score improves tremendously on NYT+FB, to the extent that it is able to beat the MFreq$(e_2|e_1^*)$ baseline by a small margin. We conclude that OOV training is essential for realizing the full potential of MF models.

\section{Joint MF-TF Models}
\label{sec:hybrid}

{\bf Background on Joint MF-TF Models:} Recall that Singh \etal\ \shortcite{singh2015towards} compare TF and MF models (particularly, E and F) and find that they have complementary strengths. In response they develop joint TF-MF models and find that they outperform individual models on artificial datasets and NYT+FB. Their best model (E+F) uses the scoring function $\phi^{E+F} = \sigma(\phi^{E}+\phi^F)$, where $\sigma$ is the sigmoid function. We call this model an \textbf{additive score (AS)} joint model, since the scores of two models are added. Early works of Reidel \etal\ \shortcite{reidel-naacl13} also experiment with a joint model for NYT+FB. Later, Toutanova \etal\ \shortcite{toutanova2015representing} implement a joint E+DM+F model and tested it on FB15K-237 but no other datasets.

We are motivated by developing a model that is robust across all datasets. 
Do additive score E+F or additive score E+DM+F meet this requirement?

\begin{table}[t]
\centering
{\small
\begin{tabular}{|l|c|c|} \hline
Dataset & $\Delta$ MRR & $\Delta$ HITS@10\\ \hline
FB15K-237 & -3.38 & -3.71 \\
FB15K & -20.48 & -27.22 \\
NYT+FB & -60.94 & -69.26 \\
WN18 & -19.17 & -18.00 \\ \hline
\end{tabular}
\caption{Change in performance of DM model initialized with corresponding embeddings extracted from DM+F (AS).}
\label{dm-extracted}
}
\end{table}

\noindent
{\bf Additive loss (AL) joint model:} Our goal is to develop one joint model that can at least match the performance of the best individual model for {\em each} dataset. We focus on joint DM+F models. 

Preliminary investigations reveal that additive score models can suffer substantial loss in performance on some datasets. Table \ref{dm-extracted} shows drop in performance in the DM component when trained jointly in additive score DM+F model. It clearly shows that DM's performance can reduce drastically due to joint training. A primary reason is that F scores overshadow DM (and E) scores.\footnote{To calibrate them, we tried standardizing scores obtained from pre-trained models. We also tried to learn a slope and bias to push DM and F model scores to the same range simultaneously. We also tried sharing of relation parameters to allow information to flow from DM to MF. Unfortunately, none of the approaches were robust across datasets.} 
Moreover, the number of parameters in MF models (vectors for entity pairs) significantly outnumber those in TF models (vectors for entities). This can lead to significant overfitting.



In response, we develop a different class of joint models in which instead of adding the scores ($\phi$s), we add their loss functions: $\mathcal{L}^{DM+F}=\mathcal{L}^{DM} + \mathcal{L}^{F}$. We name these {\em additive loss} joint models (AL).  
We expect this to be more resilient to overshadowing, since the joint loss expects each model’s individual loss to decrease as much as possible. One may note that AL style of training is equivalent to training the models separately. However, joint training makes other extensions possible, such as regularization.


\begin{table*}[!htbp]
\centering
{\small
\begin{tabular}{|l|l|cc|cc|cc|cc|} \hline
& \multirow{2}{*}{Model} & \multicolumn{2}{c|}{FB15K} & \multicolumn{2}{c|}{FB15K-237} & \multicolumn{2}{|c|}{WN18} & \multicolumn{2}{c|}{NYT+FB}   \\
& & MRR & HITS@10 & MRR & HITS@10 & MRR & HITS@10 & MRR & HITS@10 \\ \hline
\hline
1 & F  & 17.94 & 23.82 & 0.0 & 0.0 & 0.19 & 0.24 & 81.51 & 93.67  \\  
2 & DM &  44.70 & 66.26 & {\bf 34.07} & 52.93 & {\bf 75.91} & {\bf 94.12} & 62.48 & 72.17   \\ \hline
3 & E+F (AS) & 26.24 & 37.35 & 29.71 & 44.39 & 1.60 & 4.04 & {\bf 82.46} & 92.21  \\ 
4 & DM+F (AS) & 22.41 & 35.81 & 19.81 & 41.95 & 41.54 & 73.32 & 81.48 & 93.47  \\ 
5 & DM+E+F (AS) & 29.89 & 42.00 & 33.65 & 49.26 & 22.92 & 39.26 & 81.41 & 91.41 \\ \hline
6 & DM+F (AL) & 37.61 & 59.0 & 26.77 & 49.77 & 73.95 & 93.22 & 82.28 & 95.63  \\ 
7 & DM+F (RAL) & {\bf 45.81} & {\bf 67.64} &  33.38 & {\bf 53.24} & 74.55 & 93.46   & 82.28 & {\bf 95.63}  \\ \hline \hline
8 & DM+F (Oracle) & 49.42 & 69.00 & 34.07 & 52.93 & 75.95 & 94.16 & 86.06 & 95.73  \\ \hline
\end{tabular}
\caption{Performance of joint models.  AL = additive loss.  AS = additive score.
DM+F combined with regularized additive loss (RAL) is most robust across all datasets.}
\label{eval-joint}
}
\end{table*}

\paragraph*{Regularized additive loss (RAL):}
 We extend the vanilla AL joint model to a {\em regularized} joint model in which the parameters of MF model are L2-regularized. We expect this regularization to encourage a reduction in overfitting caused due to the large number of MF parameters. Overall, our final joint model has the loss function:
{\small 
\begin{equation*}
\mathcal{L}^{DM+F}(\theta^{DM}, \theta^{F}) = \mathcal{L}^{DM}(\theta^{DM}) + \mathcal{L}^{F}(\theta^{F}) + \lambda \norm{\theta^{F}}_2
\end{equation*}
}
At test time, for a query $ \langle e_1^*, r^*, ? \rangle $ an AL model cannot simply add the scores, since some entity-pairs may be OOVs. We develop various backoff cases, reminiscent of traditional backoff in language models \cite{DBLP:books/daglib/0001548}. For every $e_2$: 
\begin{itemize}
\item {\bf Case 1:} $(e_1^*,e_2) \in \mathcal{T}_{tr}$. Score of tuple is $\phi^{DM}(e_1^*, r^*, e_2) + \phi^{F}(e_1^*, r^*, e_2)$.

\item {\bf Case 2:} $(e_1^*,e_2) \notin \mathcal{T}_{tr}$, but $e_2$ is seen in training. Score of tuple is $\phi^{DM}(e_1^*, r^*, e_2) + \phi^{F}(e_{oov}, r^*, e_{oov})$. 

\item {\bf Case 3:} $e_2$ is not seen in training. Score of tuple is $\phi^{DM}(e_1^*, r^*, e_{oov}) + \phi^{F}(e_{oov}, r^*, e_{oov})$.
\end{itemize}

\begin{table}[b!]
\centering
{\small
\begin{tabular}{lcccc}
\hline
\multicolumn{1}{|l|}{\multirow{2}{*}{Model}} & \multicolumn{2}{c|}{OOV}                                  & \multicolumn{2}{c|}{Non-OOV}                              \\ 
\multicolumn{1}{|l|}{}                       & \multicolumn{1}{c}{MRR}   & \multicolumn{1}{c|}{HITS} & \multicolumn{1}{c}{MRR}   & \multicolumn{1}{c|}{HITS} \\ \hline

\multicolumn{1}{|l|}{F}                     & \multicolumn{1}{c}{0.01}  & \multicolumn{1}{c|}{0}       & \multicolumn{1}{c}{57.33} & \multicolumn{1}{c|}{75.98}   \\ 
\multicolumn{1}{|l|}{DM}                     & \multicolumn{1}{c}{36.9}  & \multicolumn{1}{c|}{58.07}   & \multicolumn{1}{c}{61.82} & \multicolumn{1}{c|}{84.25}   \\ 
\multicolumn{1}{|l|}{DM+F (AS)}                 & \multicolumn{1}{c}{14.69}  & \multicolumn{1}{c|}{29.79}   & \multicolumn{1}{c}{39.37} & \multicolumn{1}{c|}{49.04}   \\ 
\multicolumn{1}{|l|}{DM+F (RAL)}                 & \multicolumn{1}{c}{\bf 38.06} & \multicolumn{1}{c|}{\bf 59.54}   & \multicolumn{1}{c}{\bf 62.84} & \multicolumn{1}{c|}{\bf 85.42}   \\ \hline

\end{tabular}
\caption{Performance segregated by OOV and non-OOV test queries on FB15k. DM+F (RAL) matches best models for both OOV and non-OOV.}\label{oov-vs-nonoov}  }

\par \bigskip

\centering
{\small
\begin{tabular}{|l|c|c|}
\hline
\textbf{Dataset}  &   \textbf{Singleton Rate} & \textbf{Doubleton Rate} \\ \hline
FB15K               & 83.83\% & 12.19\%            \\ 
FB15K-237            & 90.52\% & 8.64\%              \\ 
WN18                   & 99.80\% & 0.20\%             \\ 
NYT+FB                & 8.06\% & 59.04\%             \\ \hline
\end{tabular}
\caption{The fraction of entity-pairs occurring exactly once and exactly twice. NYT+FB has an unusual distribution.}
\label{dataset-details}
}
\end{table}


\noindent{\bf Results:}
Table \ref{eval-joint} compares the performance of individual models with joint models.  Regularization penalty $\lambda$ is chosen over a small devset from within the training set.  All joint models are trained using both max-margin and log-likelihood losses, and we report the better of the two.

We find that different additive score models (rows 3--5) perform well on some datasets, but are not robust across them. For example, in FB15K none of these are able to match up to DM's performance. We attribute this to overfitting by F, which makes the model believe that $\phi^F$ is predicting the tuple very well.  This lets F override TF and reduces the joint model's need to learn the best TF model(s). Note that row 3 and row 5 are the models reported in \cite{singh2015towards} and \cite{toutanova2015representing}, respectively.

Rows 6 and 7 report the results of additive loss DM+F models, both without and with regularization. As anticipated, adding the losses improves performance since both models get trained well. Moreover, regularization also helps considerably since now the model is not overwhelmed by too many F parameters. RAL version of DM+F achieves scores close to the best individual model on each dataset.  In some cases, its performance is marginally weaker, and in other cases it is slightly better. Overall, this model has the desired robustness across datasets. 



\noindent{\bf Analysis: }
Row 8 of Table~\ref{eval-joint} also shows the accuracy of an oracle model that, for every test query, post-facto selects the model with the more accurate score (between DM and F). This upper bounds the performance expected from a perfect joint DM+F model, fixing the constituents. We find that the oracle is only 3-4 MRR percentage points better than our best model for two datasets, and the differences are much less for the other two.  Overall, it suggests that our proposed joint model obtains a strong robust performance.

Table \ref{oov-vs-nonoov} breaks down the performance of models on the subset of test queries that have OOVs and non-OOV gold entity pairs. This analysis is meaningful only for FB15K, since other datasets have extreme entity-pair OOV rates (see Table \ref{dataset-oovs}). We observe that while F has extremely poor performance on OOVs (and thus weak performance overall), it performs decently on non-OOVs. RAL DM+F is able to perform well on both OOVs and non-OOVs, whereas DM+F (AS) has poorer performance on both of them (although still better than vanilla F for OOVs). Also note that F is outperformed by DM even on non-OOVs; this refutes prior claims that F always performs better than TF models when test entity pairs are seen during training \cite{reidel-naacl13,toutanova2015representing}.

\section{Discussion and Future Work}
We now list two observations that suggest important directions for future research in KB inference. 

\paragraph{Dataset Characteristics:} Our work subjects datasets to natural sanity checks. First, we introduce two most frequent baselines (Table \ref{eval-scores}) to understand the nature of the KBs. Second, we compute entity-pair OOV rates (Table~\ref{dataset-oovs}) as a rough predictor of the relative success of the TF and MF families. Finally, in Table \ref{dataset-details}, we report the singleton and doubleton percentages (for entity pairs). A singleton is an entity-pair occurring only once in the data ($\mathcal{T}_{tr} \cup \mathcal{T}_{ts}$) and a doubleton is an entity pair that occurs exactly twice. Doubletons have a strong effect in the scenario painted in Table~\ref{eval-example}. We find that almost every dataset has some idiosyncrasy, which raises the question whether it is a good representative for the datasets found naturally.

In particular, WN-18 and FB15K-237 have near 100\% entity-pair OOV rates, unlikely to be the case in real KBs. In FB15K-237 the best models are not much better than MFreq($e_2|r^*$) baseline. This is because the dataset is artificially constructed to avoid relations with entity-pair overlap. But, this reduces its ability to make many interesting inferences. For NYT+FB, MFreq($e_2|e_1^*$) performance has a strong performance with 95\% score on HITS@10. Moreover, learned models are able to improve its MRR by only about three percentage points. Statistics in Table \ref{dataset-details} reveal that this could be because the dataset has an unusually high number of entity-pair doubletons: it is the only data set where doubletons by far outnumber singletons. It is unlikely that such a distribution occurs in a naturally occurring dataset. FB15K appears to pass our sanity tests. We believe that focus on better datasets will likely help us in better progress on KB inference.

\paragraph{Path based inference:}
In KBs, a common type of inference is based on relation paths (or Horn-clauses), e.g., (Michael Jordan, teaches at, Berkeley) and (Berkeley, is located in, California) implies (Michael Jordan, teaches in, California). To assess the ability of inference models to automatically learn such relation paths, we tested them on artificial datasets, where we provided many instances of two-hop paths with relations $r_1$ and $r_2$ implying a third relation $r_3$. We find that {\em none} of the four models are effective at predicting such relations. A study similar to ours comparing the latest models that train over relation paths \cite{guu2015traversing,DBLP:conf/emnlp/Garcia-DuranBU15,toutanova2016compositional} will benefit our understanding of path-based inference.

\section{Related Work}
\label{sec:related}

Traditional methods for inference over KBs include random walks over knowledge graphs \cite{lao2011random}, natural logic inference \cite{maccartney2007natural}, and use of statistical relational learning models such as Markov Logic Network, Bayesian Logic Programs, and Probabilistic Soft Logic \cite{schoenmackers2008scaling,raghavan2012learning,yangwang2015joint}. These need (or benefit from) a background knowledge of inference rules, predominantly generated via extended distributional similarity \cite{lin2001dirt,schoenmackers2010learning,nakashole2012patty,galarraga2013amie,grycner-EtAl:2015:EMNLP,berant-acl12,DBLP:conf/naacl/JainM16}.

Neural methods for KB inference combine both inference and rule learning into one unified framework to add new facts to the KB directly. Both MF and TF methods have been very popular with several extensions proposed for each. The original F model has been extended to incorporate first order logic rules, \cite{rocktaschel-naacl15,demeesterregularizing}, to predict for relations not seen at training time \cite{verga2015multilingual}, etc. It has also been extended to generate embedding of a new entity-pair on the fly \cite{verga2016generalizing}. But that is different from our OOV method, since, at test time, they expect knowledge of several tuples between the same entity pair.

Similarly, other TF models also exist, for example, Parafac \cite{harshman1970foundations}, Rescal \cite{nickel2011three} and NTN \cite{socher2013reasoning}. These are older models which are shown to be outperformed by models evaluated in this paper.  More recent models have also been introduced such as a model using holographic embeddings \cite{nickel2015holographic}, and another with asymmetric embeddings using complex vectors \cite{trouillon2016complex}. It will be nice to compare these rigorously as well. The learned embeddings can use additional information such as typing \cite{DBLP:conf/emnlp/ChangYYM14}, have been used to mine logical rules \cite{yang15export:241703} and have been used for schema induction \cite{nimishakavi2016relation}.

\section{Conclusion}


We extensively evaluate various tensor factorization (TF) and matrix factorization (MF) models for KB inference on all popular datasets. After replacing the standard evaluation protocol with our proposed OOV-cognizant KBI protocol, we find that DistMult (a TF model) is fairly robust across a variety of datasets, but F (an MF model) outperforms others on one dataset. F's performance increases further by training an OOV entity-pair vector. Finally, we propose joint models that combine DistMult and F. We find that adding the loss functions from both models with a regularization on F's parameters achieves the most robust results across all datasets. 

We also present a series of analyses of our empirical results. First, our work increases our understanding of relative strengths and weaknesses of MF and TF models given some important bulk characteristics of the data sets. Specifically, we establish a strong connection between accuracy of various approaches and the fraction of OOV test entity pairs, and the proportion between entity pair singletons and doubletons. Second, we find that our joint model achieves results at par with the best individual models for both OOV and non-OOV queries. As a by-product, we identify some peculiarities in existing datasets, which suggests a need to design better benchmark datasets.
 
We release our code for all models and evaluation protocols for further use by research community. In the future, we wish to study models that explicitly incorporate relation paths for KB inference.

\bibliography{acl2012}
\bibliographystyle{acl2012}

\end{document}